\documentclass[10pt, twocolumn]{article} 

\usepackage[numbers]{natbib} 
\usepackage{graphicx}       
\usepackage[utf8]{inputenc} 
\usepackage{tabularx}       
\usepackage{booktabs}       
\usepackage{longtable}      
\usepackage{array}          
\usepackage[table,xcdraw]{xcolor} 
\usepackage{tcolorbox}      
\usepackage[margin=1in, top=0.8in]{geometry} 
\usepackage{enumitem}       
\usepackage{multirow}       
\usepackage{adjustbox}      
\usepackage{caption}        
\usepackage{float}          
\usepackage[sfdefault]{FiraSans} 
\usepackage{multicol}       
\usepackage{parskip}
\usepackage{ragged2e}
\usepackage{appendix}
\usepackage{siunitx}
\usepackage{microtype}
\newcolumntype{P}[1]{>{\raggedright\arraybackslash}p{#1}}
\usepackage{titlesec}
\titleformat{\section}{\large\bfseries}{\thesection.}{1em}{}
\titleformat{\subsection}{\normalsize\bfseries}{\thesubsection.}{1em}{}
\titleformat{\subsubsection}{\small\bfseries}{\thesubsubsection.}{1em}{}

\usepackage{fancyhdr}
\renewenvironment{abstract}
  {\noindent\textbf{\abstractname}\par\noindent\ignorespaces}
  {\par\noindent\ignorespacesafterend}

\title{\vspace{-1cm}\textbf{\Large A Scalable Approach to Benchmarking the In-Conversation Differential Diagnostic Accuracy of a Health AI}} 
\author{
Deep Bhatt, Surya Ayyagari\, Anuruddh Mishra\thanks{Corresponding author: \href{mailto:anuruddh@getbeyondhealth.com}{anuruddh@getbeyondhealth.com}} \\
\small \textsuperscript{1}Medical Evaluation, August AI, Bengaluru, Karnataka, India \\
}

\date{}
\usepackage{hyperref}       
\begin{document}

\bibliography{main.bib}

\twocolumn[
\begin{@twocolumnfalse}
\maketitle
\vspace{-0.5cm} 
\begin{abstract}
\noindent
Diagnostic errors in healthcare persist as a critical challenge, with increasing numbers of patients turning to online resources for health information. While AI-powered healthcare chatbots show promise, there exists no standardized and scalable framework for evaluating their diagnostic capabilities. This study introduces a scalable benchmarking methodology for assessing health AI systems and demonstrates its application through \textit{August}, an AI-driven conversational chatbot. Our methodology employs 400 validated clinical vignettes across 14 medical specialties, using AI-powered patient actors to simulate realistic clinical interactions. In systematic testing, \textit{August} achieved a top-one diagnostic accuracy of \textbf{81.8\%} (327/400 cases) and a top-two accuracy of \textbf{85.0\%} (340/400 cases), significantly outperforming traditional symptom checkers. The system demonstrated \textbf{95.8\%} accuracy in specialist referrals and required 47\% fewer questions compared to conventional symptom checkers (mean 16 vs 29 questions), while maintaining empathetic dialogue throughout consultations. These findings demonstrate the potential of AI chatbots to enhance healthcare delivery, though implementation challenges remain regarding real-world validation and integration of objective clinical data. This research provides a reproducible framework for evaluating healthcare AI systems, contributing to the responsible development and deployment of AI in clinical settings. 
\end{abstract}
\vspace{0.5cm}
\end{@twocolumnfalse}
]

\clearpage
\section{Introduction}
Clinical decision making as the cornerstone of medical practice is heavily based on accurate diagnoses to guide patient care~\citep{ref1, ref2}. Diagnostic errors, which can lead to significant patient harm, arises from a blend of systemic and cognitive factors~\citep{ref3, ref4, ref5, ref6, ref7}. As patients increasingly rely on digital resources for health information, the landscape is evolving. A 2010 survey involving more than 12,000 individuals across 12 countries found that 75\% of respondents sought health information online~\citep{ref8}. In 2017, approximately 67\% of patients admitted to searching for their symptoms on Google before visiting a doctor~\cite {ref9}. More than 33\% of adults in the United States regularly use the internet to self-diagnose their ailments, using it both for non-urgent symptoms and urgent symptoms such as chest pain~\citep{ref10}. It's not just limited to primary healthcare either, with a recent study revealing that 50\% of patients researched their symptoms online before going to emergency departments~\cite {ref11, ref12}.

The evaluation of healthcare AI systems presents unique challenges that traditional medical assess-ment frameworks struggle to address ~\cite{ref64} ~\cite{ref65}. Current evaluation methods typically rely on three approaches: multiple-choice medical questions, structured clinical vignettes with predefined responses, or limited human-based testing. Each of these methods has significant limitations in assessing AI systems designed for patient interaction. Multiple-choice question datasets like USMLE MedQA evaluate only the final diagnostic decision rather than the critical process of information gathering. 

This approach fails to assess the AI's ability to conduct appropriate clinical reasoning or adapt to varying patient presentations. Human-based testing, while valuable for assessing real-world performance, faces significant scalability constraints and potential inconsistencies in how actors present symptoms or respond to questions~.

Traditional clinical vignette assessments often employ rigid response structures that fail to capture the dynamic nature of patient-provider interactions. These assessments typically focus on the presentation of symptoms rather than the interactive process of history-taking, which is crucial for accurate diagnosis in clinical practice. This approach also runs the risk of the exact or similar vignette being present in the AIs training data.
Moreover, existing benchmarking approaches often overlook critical aspects of healthcare delivery such as:
\begin{itemize}
    \item The ability to handle ambiguous or contradictory patient information
    \item The management of varying levels of health literacy and communication styles
    \item The capacity to maintain consistency while adapting to individual patient needs
    \item The effectiveness of gathering pertinent negative findings through conversation
\end{itemize}

This research gap necessitates a new approach to benchmarking healthcare AI systems - one that combines the standardization of clinical vignettes with the interactive nature of real patient encounters, while maintaining scalability for practical implementation. Our proposed framework addresses these limitations by introducing AI-powered patient actors that can consistently simulate realistic patient interactions while allowing for standardized evaluation across a large number of clinical scenarios.

Relying on the internet has broadly become the norm. This shift highlights both an opportunity allowing for the rise of unique online conversational AI platforms which can aid patients as an efficient and accessible tool for seeking online health information~\citep{ref13}. However, this opportunity highlights a challenge:  while digital tools can empower patients, they also risk disseminating inaccurate information. 

\subsection{Traditional Symptom Checker Systems}

Traditional history-taking—a critical component of diagnosis—involves a nuanced dialogue between clinician and patient, capturing not just symptoms but also the contextual factors that influence health. However, time constraints and resource limitations can hinder this process. In response, digital health technologies have emerged, including online symptom checkers and, more recently, AI-powered chatbots designed to assist with differential diagnosis. Despite their potential, many of the former lack the conversational ability to replicate the depth of a clinical interview. They often rely on checklists, simple queries, or generative adversarial networks to provide a diagnosis, failing to capture the subtleties of patient narratives.

\begin{figure}[h!]
    \centering
    \includegraphics[scale=0.5]{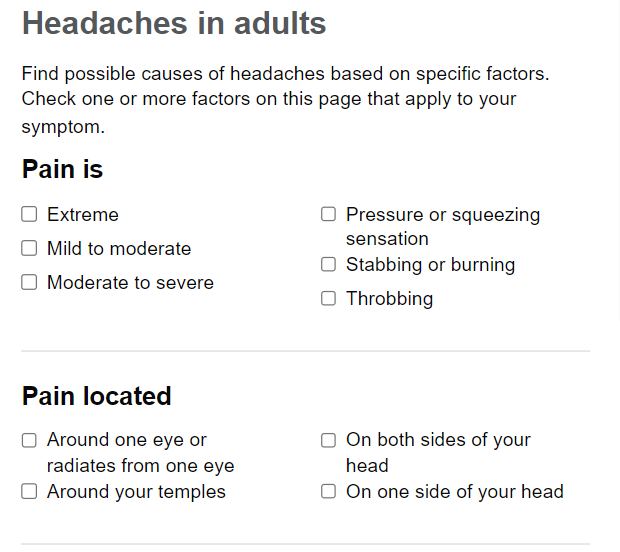} 
    \caption{Mayo Clinic's Symptom Checker.}
\end{figure}

Online symptom checkers typically require users to input symptoms using specific terminology that aligns with the designers' expectations. This rigid input method often means that if users describe their symptoms differently—perhaps using layman's terms, vaguely describing a sequence of symptoms or expressing multiple symptoms simultaneously — the system may fail to interpret their condition accurately. Such tools are not conversational and lack the capability for back-and-forth interactions that clarify ambiguities or explore the user’s concerns further. This absence of a conversational interface results in a suboptimal user experience (UX, highlighted in Fig.1), as users may feel misunderstood or frustrated by the system’s limitations~\citep{ref51}. This results in varying degrees of diagnostic accuracy ranging from 5-50\% for top-1 accuracy, and a maximum of 95\% only when measured across 20 responses. This inaccuracy in the results of symptom checkers does not provide the due diligence required by healthcare care, jeopardizing the well-being of patients. ~\citep{ref36}

Moreover, traditional symptom checkers generally do not offer recommendations for the next steps or clarify doubts. They might provide a list of possible conditions but leave users uncertain about what actions to take. This lack of guidance can lead to anxiety or, conversely, complacency if users underestimate the seriousness of their symptoms. Additionally, these tools are ineffective at processing subjective information such as the user’s emotional state, personal concerns, or how symptoms impact their daily life. Such subjective details are crucial for a comprehensive assessment and for initiating a therapeutic relationship~\citep{ref52}.

\subsection{Challenges in Traditional Healthcare Systems}
The traditional healthcare system often struggles to provide personalized care. Due to time constraints and heavy patient loads, doctors frequently allocate less than five minutes per consultation in regions that encompass over half the global population, contributing to professional burnout~\citep{ref61,ref62}. This limited time frame hinders the thorough study of the history, resulting in missed information and rushed appointments that prioritize immediate symptoms over underlying causes. In addition, access to healthcare presents a significant challenge, particularly in underserved areas where medical professionals are in short supply. Even when care is accessible, patients often face difficulties reaching the appropriate specialist promptly due to long wait times, complex referral processes, social, economical and geographic barriers.

\subsection{Rise of AI-powered Chatbots}

About one in six adults (17\%) say they use AI chatbots at least once a month to find health information and advice, rising to one-quarter of adults under age 30 (25\%)~\citep{ref15}, and most physicians anticipate a significant role for such AI tools in future clinical practice~\citep{ref45}. Moreover, concerns about the accuracy, reliability, and safety of AI chatbots persist, compounded by a lack of standardized benchmarks to evaluate their performance in clinical settings~\cite {ref16, ref17, ref18, ref19}. We present a comprehensive benchmarking framework to study the diagnostic ability of a conversational AI chatbot designed for differential diagnosis. We evaluate the chatbot's ability to engage users in natural language, accurately interpret patient-reported symptoms, and provide reliable diagnostic suggestions. By assessing the chatbot's performance against established clinical standards, we aim to determine its viability as a tool to support both patients and clinicians in the diagnostic process.

The fundamental underpinning of these chatbots are  Large Language Models. Large language models (LLMs) are machine learning systems that can generate human-like responses from written language inputs~\citep{ref23, ref24, ref25}. These LLMs have demonstrated capabilities that are relevant to healthcare, such as the ability to solve complex medical cases, exhibit human-like clinical reasoning, take patient histories, and display empathetic communication~\citep{ref26, ref27, ref28, ref38, ref40}. Studies have shown that LLMs can be prompted to mimic the diagnostic reasoning processes of clinicians while maintaining accuracy, addressing concerns about the interpretability of large language models in medicine and enhancing their potential for safe patient care~\citep{ref35}. LLMs have demonstrated superiority in providing appropriate and interactive disease prevention advice, effectively supporting clinical workflows and patient communication~\citep{ref47}.

Integrating AI chatbots into healthcare holds promise for enhancing patient engagement, increasing access to medical guidance, and alleviating some burdens on healthcare systems~\citep{ref20, ref21, ref22}. Conversational diagnostic AI offers a more personalized assessment for self-triage and diagnosis compared to traditional Google searches, thereby enhancing user experience~\citep{ref48}. However, rigorous evaluation is essential to ensure these technologies are safe and effective contributing positively to patient outcomes. This study contributes to the body of knowledge necessary for responsibly integrating  AI into clinical decision-making.

\section{Background}
The utility and promise of conversational diagnostic chatbots cannot be realized if they are not accurate in initial symptom checking, triaging, and differential diagnosis. There are several limitations to the current approach of benchmarking chatbots against publicly available medical multiple-choice question datasets like the USMLE MedQA. The multiple-choice format inherently makes it easier for models to select the correct option rather than formulating a diagnosis from scratch, which does not adequately assess the model's ability to synthesize information and engage in clinical reasoning~\cite {ref29}. These benchmarking standards are further thrown into question when considering the diversity of input data, i.e. the complement data processed through diagnostic reports and other such lab diagnostics which assist a doctor in their decision-making. 

A major area where conversational diagnostic chatbots surpass conventional online symptom checkers is in naturally gathering relevant clinical information from the patient~\cite {ref30}. A significant part of clinical reasoning involves history-taking and being able to ask open-ended questions—both leading and non-leading—to arrive at a diagnosis~\citep{ref31, ref32, ref33}. This critical skill set remains untested in multiple-choice question-based datasets.

Moreover, most of these datasets do not accurately resemble real-world clinical scenarios. They typically include only classic presentations and typical findings, lacking the complexity, ambiguity, and atypical cases that healthcare professionals encounter in practice. This means they fail to capture the nuanced decision-making required in actual medical settings~\citep{ref37, ref39}. Consequently, this approach may not effectively evaluate a chatbot's practical utility or its readiness to handle the uncertainties inherent in medical practice.

Clinical vignettes are typically used to test physicians and other clinicians on their diagnostic ability and management decisions in life-like scenarios. After the seminal study by Semigran et al., they became the de facto benchmark for quantifying the performance of online symptom checkers~\citep{ref36}. Simply giving clinical vignettes to the AI to diagnose does not accurately simulate how the AI will perform while actually communicating with the user. The framework we propose relies on creating accurate simulated patient actors which can hold a clinical diagnostic interaction with the health AI that is similar to the way an actual patient would interact. We leverage validated clinical vignettes as input for the patient actor. Using simulated patient actors versus actual humans allows us to rapidly scale and iterate the accuracy of the health AI across different factors such as disease, age, sex, and race. While this is an important step forward from existing benchmarking methods, extensive real-world health benchmarks will still be required before a health AI is deployed in a clinical setting.

We applied the proposed framework to compare \textit{August}.  August, a health AI, thoroughly researched and iterated specifically to help people access high-quality health information. Since its launch one year ago, \textit{August} has gained over 1.5 million users across 160 countries, thus validating the quality of August's overall performance. We leverage this framework to assess \textit{August}'s top one and top two diagnostic suggestions with the reference diagnoses provided in the original clinical vignettes. A total of 400 clinical vignettes were selected from an established database to represent a wide spectrum of medical conditions across different specialties~\citep{ref50}. This comparison aims to evaluate \textit{August}'s ability to generate accurate diagnostic hypotheses based solely on the information elicited from patient-reported symptoms.

\section{Benchmark Design}

“The benchmark we are proposing consists of three distinct steps. The process begins with the creation of AI-based \textit{patient actors} based on clinical vignettes that represent cases one might encounter in a hospital. This is followed by an extensive interaction with the Health AI, which actively identifies possible symptoms and provides its diagnosis. Lastly, conversation moderators and judges review the interactions, enabling scalable testing and assessment of the Health AI's accuracy 

\begin{figure}[h!]
    \centering
    \includegraphics[scale=0.25]{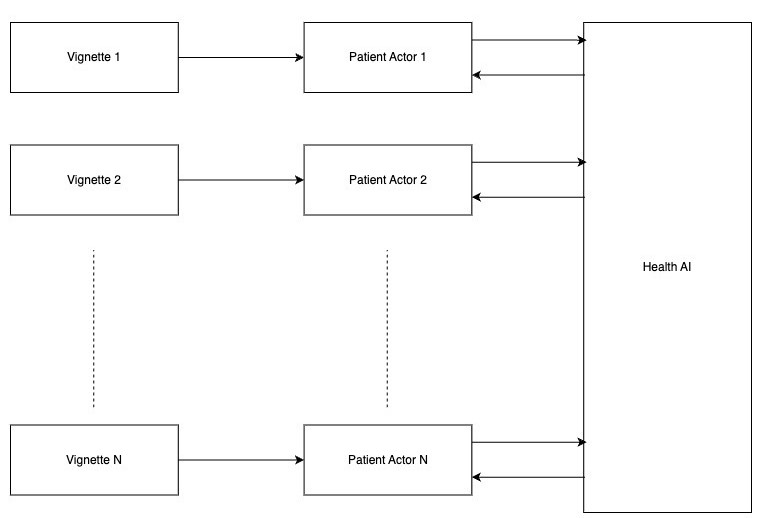} 
    \caption{Schematic of an Interaction}
\end{figure}

This proposed simulation-based benchmark is later utilized to evaluate the diagnostic accuracy of \textit{August}, an AI expert system, in interpreting patient-reported symptoms. The primary objective of the benchmark is to put forward a scalable, low-cost way to assess the diagnostic accuracy of health AIs. It assess the AIs information-gathering ability through conversation and its capacity to generate accurate diagnostic hypotheses based solely on subjective patient information, mimicking initial consultations or remote telehealth encounters.

\subsection{Selection of Clinical Vignettes}
The selection of representative clinical vignettes is crucial for assessing the accuracy of differential diagnostic conversations across diverse scenarios, including variations in disease presentation and socioeconomic factors. By utilizing a broad range of clinical vignettes, we can ensure that healthcare AI systems are both accurate and inclusive. Since one of the most pressing use cases for healthcare AI is providing high-quality health information to underserved populations worldwide, ensuring an accurate representation of these communities is essential for the AI to fulfill its potential. Our framework for selecting clinical vignettes currently considers the following factors:

\textbf{Specialty}
\begin{itemize}
    \item Cardiovascular
    \item Dermatology
    \item Endocrine
    \item ENT
    \item Gastroenterology
    \item Hematology
    \item Infectious diseases
    \item Nephrology
    \item Neurology
    \item Obstetrics and Gynecology
    \item Ophthalmology
    \item Orthopedics and Rheumatology
    \item Respiratory
    \item Urology
\end{itemize}
\vspace{1em}
\textbf{Disease Diversity}
\begin{itemize}
    \item Common
    \item Uncommon
    \item Acute
    \item Chronic
\end{itemize}
\vspace{1em}
\textbf{Presentation Diversity}
\begin{itemize}
    \item Typical
    \item Atypical
    \item Uncommon
\end{itemize}
\vspace{1em}
\textbf{Socio-economic Diversity}
\begin{itemize}
    \item Age
    \item Sex
    \item Ethnicity
    \item Location
    \item Language
    \item Language fluency
\end{itemize}
\vspace{1em}
In our benchmarking of \textit{August}, we implemented a scientific methodology using the standard clinical vignette approach established by Hammoud et al~\citep{ref63}. We selected 400 clinical vignettes from their work, chosen specifically for their diversity and representation of various medical conditions. These vignettes span a wide spectrum of diseases across multiple specialities, incorporating common, uncommon, acute, and chronic conditions to thoroughly evaluate the AI system's diagnostic capabilities. The standardized scenarios ensured consistent information presentation to \textit{August} throughout the evaluation. As part of our ongoing commitment to inclusivity, we are currently collecting additional vignettes that represent a diverse cross-section of the population.

\subsection{Patient-Actor Design}
The patient actor is an LLM-based AI system built around a clinical vignette, designed to simulate conversations as a real patient would. The clinical vignette incorporates essential patient characteristics including age, sex, and race, with the patient actor programmed to reflect these factors in its conversational style and responses. Drawing from our internal research, we incorporated patterns from anonymized conversations between a health AI and actual users to inform the patient-actor's communication style. The AI patient-actor was then precisely prompted to follow these communication patterns, ensuring consistent and realistic patient interactions with the health AI. The guidelines we established for the AI patient-actor are as follows:

\begin{itemize}
    \item Type in simple language, the way a patient would, avoiding medical jargon.
    \item Present the most distressing symptom first
    \item Answer only when asked and do not volunteer extra information.
    \item Stick strictly to the vignette provided and maintain consistency throughout.
    \item Do not ask questions at the beginning but introduce any patient questions naturally later in the conversation.
    \item If the vignette is of a person \textless{18} or \textgreater{80} then talk as a proxy rather than the first person. 
    \item Keep the instructions confidential and do not mention the vignette or that you're following a script.
\end{itemize}

The outputs from the patient-actor were then reviewed and iterated on to ensure that it always adhered to the above rules. During the final evaluation of the benchmark, we thoroughly reviewed the messages from the patient-actor to ensure these guidelines were followed throughout each conversation. A simple interaction between our patient-actor and the expert is presented in Appendix A. 

Most similar studies have used human patient-actors. Our decision to use an AI patient-actor is crucial to ensuring that this benchmarking framework is scalable as well as feasible to be run regularly. AI patient-actor allows us to more accurately represent real-world conversations from patients of diverse socio-economic backgrounds.

\subsection{Judge}
The process of evaluating the accuracy of the output of the health AI is crucial for establishing the reliability of the AI. Several factors complicate the evaluation of its output. The AI, much like various human doctors, may use different names to refer to the same disease described in the vignette. In addition to this, the AI might also identify a more or less specific version of the disease the patient actually has. For our benchmark, we adopted the matching criteria from Gilbert et al., as also utilized in Hammoud et al., with certain modifications to conduct a more stringent assessment.
\vspace{1em}

A condition was considered a match when:
\begin{enumerate}
    \item \textbf{Exact Correspondence}: The predicted condition is identical to that in the gold standard vignette.
    \item \textbf{Alternative Terminology}: Alternative names for the same condition are used. For example, \textit{shoulder impingement syndrome} and \textit{rotator cuff tendinitis} refer to the same condition.
    \item \textbf{Increased Specificity}: The predicted condition is more specific or detailed than the gold standard condition. For instance, if the gold standard diagnosis is \textit{adrenal insufficiency} and the predicted diagnosis is \textit{Addison’s disease}, it is considered a match.
    \item \textbf{Equivalent Descriptions by Different Physicians}: It is reasonable to assume that two different doctors might use different descriptions to label the same condition. For example, diagnosing \textit{heart failure and pulmonary congestion due to valvular heart disease following a history of rheumatic fever} is matched with the gold standard condition \textit{mitral stenosis}
    \item \textbf{Direct Causation}: One condition directly and explicitly causes the other. For example, if the gold standard condition is \textit{Urethritis} and the predicted condition is \textit{Gonorrhoea}, it is considered a match because \textit{Gonorrhoea} directly causes urethritis.
\end{enumerate}

The following criteria from Gilbert et al. were not considered a match in our study to ensure a stricter assessment:
\begin{enumerate}
\item \textbf{Near Matches Due to Less Precision}: The predicted condition conveys the nature of the gold standard condition and is reasonably related but is less precise. We did not consider a condition a match even if the similarity is clear enough that in primary care practice, the conditions might be considered a near match. For example, if the gold standard condition is \textit{Cerebral Stroke} and the predicted condition is \textit{Transient Ischemic Attack (TIA)}, it is not considered a match because not all TIAs result in a Cerebral Stroke.
\item \textbf{Umbrella Terms}: The predicted condition is an umbrella term that includes the gold standard condition. For instance, if the gold standard condition is \textit{Chronic Bronchitis} and the predicted condition is \textit{Chronic Obstructive Pulmonary Disease (COPD)}, it is not considered a match, even though the symptomatic history elicited is similar.
\item \textbf{High Symptomatic Overlap}: The predicted condition is highly related to the gold standard condition and shares a significant degree of symptoms. For example, if the gold standard condition is \textit{Brain Abscess} and the predicted condition is \textit{Meningitis}, they are not considered a match even though their symptoms may highly overlap, they are distinct conditions and not directly equivalent.
\end{enumerate}

By refining these criteria, we aimed to conduct a more rigorous evaluation of diagnostic accuracy, ensuring that only precise and directly related conditions were considered matches in our study.

For the scope of the current benchmark, qualified human doctors evaluated the final output from the Health AI. In the future, we aim to build automated judges that can accurately apply the above criteria to judge the output from the health AI. This step will be crucial to ensure we can run the benchmark across thousands or millions of vignettes.

The outcome measures that we want to evaluate in this benchmark are broadly bucketed into three categories, Accuracy, Real-World Impact, and User Experience, with the accuracy of diagnosis being the most important. To quantify the outcomes, the judges reviewed the 'Patient Actor-Health AI' interactions to identify Health AI messages that explicitly mentioned the diagnosis. Once these messages were flagged, they were compared with the diagnoses provided in the gold-standard vignettes, and accuracy was evaluated.

\section{Outcome Measures}

While the accuracy of diagnosis is the primary benchmark metric that we are focusing on, we have also defined two other outcome categories, Real-World Impact, and User Experience to better encapsulate the results. Overall, we aim examine Accuracy, Real-World Impact, and User Experience keeping in mind the overall objective of providing the user an accurate, empathetic experience. For this paper, we've proposed a few high-impact measures in each category and have also evaluated August against the same.

\subsection{Accuracy}

We only looked at the top 1 and 2 diagnostic accuracy. While we recognize that top 5 diagnostic accuracy is a standard measure that many research papers and symptom checkers use, we strongly argue against top-3 and top-5 presenting two-fold reasons relying on academia and information theory. Our first argument is in concurrence with Semigran et al. ~\citep{ref36} who observe the clinical utility dramatically decreasing beyond the top two suggestions. The first two diagnoses typically capture the most likely and clinically relevant conditions. We also argue on a diminishing marginal value of additional diagnostic suggestions due to the decreasing entropy on multiple tests. With this in mind, we aim to continue evaluating just the top 1 and top 2 diagnostic accuracy.

We define these accuracies as follows: 

\begin{itemize}
\item \textbf{Top-One Diagnosis Accuracy}: Whether the Health AI's first diagnosis matched directly with the gold-standard diagnosis.
\item \textbf{Top-Two Diagnoses Accuracy}: Whether either of the Health AI's two suggestions matched the reference diagnosis.
\end{itemize}

\subsection{Real World Impact}
\begin{itemize}
\item \textbf{Correct Specialty Identified}: Would the user have reached the right specialist based on the output given by the Health AI? This accuracy is important since the goal of the health AI should not be to diagnose, but instead to ensure people get the right care they need at the right time. 
\end{itemize}

\subsection{User Experience}
Our focus here lies on the \textbf{Consultation Length}, defined as the total interaction time between the Health AI and the patient. We aim to minimize the number of questions posed by the AI based on each specialty.

\section{Benchmark Results}
\subsection{Diagnostic Accuracy}
\textit{August's} Top-1 diagnostic accuracy was 81.8\%, surpassing the 67.5\% and 54.2\% achieved by popular online symptom checkers Avey and Ada Health as shown in table 2. We also extend our measurement, using this metric to compare against expert clinicians of different years of expertise. \textit{August} outperforms them as well, exceeding their accuracies of 49.7\%, 61.3\%, and 72.5\% . One of those was a family medicine doctor with \textgreater{30} years of experience. The other 2 are also family medicine doctors, each with \textgreater{10} years of experience. Notably, \textit{August} also outperforms average primary care clinicians in the vignette setting by 34.1\%.

\sisetup{
    detect-all,
    round-mode = places,
    round-precision = 1,
    output-decimal-marker = {.},
    group-separator = {,},
    group-minimum-digits = 4,
}


\begin{table}[htbp]
\centering
\caption{Top 1 Diagnostic Accuracy by System}
\label{tab:top1_diagnostic_accuracy}
\begin{tabularx}{\columnwidth}{@{}>{\raggedright\arraybackslash}p{0.5\columnwidth}
                                >{\centering\arraybackslash}X
                                >{\centering\arraybackslash}X@{}}
\toprule
\textbf{System} & \textbf{Top 1 Correct} & \textbf{Top 1 \%} \\
\midrule
Cardiovascular            & 36 & 80.0\% \\
Dermatology               & 13 & 100.0\% \\
Endocrine                 & 14 & 77.8\% \\
ENT                       & 22 & 88.0\% \\
GI                        & 36 & 83.7\% \\
Hematology                & 16 & 69.6\% \\
Infectious                & 20 & 69.0\% \\
Nephrology                & 13 & 81.3\% \\
Neurology                 & 17 & 81.0\% \\
Obstetrics and Gynecology & 44 & 84.6\% \\
Ophthalmology             & 14 & 77.8\% \\
Orthopedics and Rheumatology & 25 & 78.1\% \\
Respiratory               & 31 & 86.1\% \\
Urology                   & 26 & 89.7\% \\
\midrule
\textbf{Grand Total}      & \textbf{327} & \textbf{81.8\%} \\
\bottomrule
\end{tabularx}
\end{table}

\begin{table}[htbp]
\centering
\caption{Top 2 Diagnostic Accuracy by System}
\label{tab:top2_diagnostic_accuracy}
\begin{tabularx}{\columnwidth}{@{}>{\raggedright\arraybackslash}p{0.5\columnwidth}
                                >{\centering\arraybackslash}X
                                >{\centering\arraybackslash}X@{}}
\toprule
\textbf{System} & \textbf{Top 2 Correct} & \textbf{Top 2 \%} \\
\midrule
Cardiovascular            & 37 & 82.2\% \\
Dermatology               & 13 & 100.0\% \\
Endocrine                 & 14 & 77.8\% \\
ENT                       & 23 & 92.0\% \\
GI                        & 39 & 90.7\% \\
Hematology                & 16 & 69.6\% \\
Infectious                & 22 & 75.9\% \\
Nephrology                & 13 & 81.3\% \\
Neurology                 & 17 & 81.0\% \\
Obstetrics and Gynecology & 47 & 90.4\% \\
Ophthalmology             & 16 & 88.9\% \\
Orthopedics and Rheumatology & 25 & 78.1\% \\
Respiratory               & 32 & 88.9\% \\
Urology                   & 26 & 89.7\% \\
\midrule
\textbf{Grand Total}      & \textbf{340} & \textbf{85.0\%} \\
\bottomrule
\end{tabularx}
\end{table}

\subsection{Real World Impact}
\textit{August}'s accuracy in referring users to the right specialist was 95.8\%. For common cases, this accuracy was 98.6\%, while for less common cases, the accuracy was 92.1\%, as shown in Tables~\ref{tab:specialist_accuracy} and~\ref{tab:incidence_accuracy}.

\begin{table}[htbp]
\centering
\caption{Specialist Recommendation Accuracy by System}
\label{tab:specialist_accuracy}
\begin{tabularx}{\columnwidth}{@{}>{\raggedright\arraybackslash}p{0.5\columnwidth}
                                >{\centering\arraybackslash}X
                                >{\centering\arraybackslash}X
                                >{\centering\arraybackslash}X@{}}
\toprule
\textbf{System} & \textbf{Total Cases} & \textbf{Correct Referral} & \textbf{\% Correct} \\
\midrule
Cardiovascular            & 45 & 42 & 93.3\% \\
Dermatology               & 13 & 13 & 100.0\% \\
Endocrine                 & 18 & 17 & 94.4\% \\
ENT                       & 25 & 25 & 100.0\% \\
GI                        & 43 & 43 & 100.0\% \\
Hematology                & 23 & 17 & 73.9\% \\
Infectious                & 29 & 28 & 96.6\% \\
Nephrology                & 16 & 14 & 87.5\% \\
Neurology                 & 21 & 21 & 100.0\% \\
Obstetrics and Gynecology & 52 & 51 & 98.1\% \\
Ophthalmology             & 18 & 18 & 100.0\% \\
Orthopedics and Rheumatology & 32 & 30 & 93.8\% \\
Respiratory               & 36 & 35 & 97.2\% \\
Urology                   & 29 & 29 & 100.0\% \\
\midrule
\textbf{Grand Total}      & \textbf{400} & \textbf{383} & \textbf{95.8\%} \\
\bottomrule
\end{tabularx}
\end{table}

\begin{table}[htbp]
\centering
\caption{Referral Accuracy by Incidence}
\label{tab:incidence_accuracy}
\begin{tabularx}{\columnwidth}{@{}>{\raggedright\arraybackslash}X
                                >{\centering\arraybackslash}X
                                >{\centering\arraybackslash}X
                                >{\centering\arraybackslash}X@{}}
\toprule
\textbf{Incidence} & \textbf{Top 1 \%} & \textbf{Top 2 \%} & \textbf{Referral Accuracy} \\
\midrule
Common      & 87.4\% & 91.0\% & 98.6\% \\
Less Common & 74.7\% & 77.5\% & 92.1\% \\
\bottomrule
\end{tabularx}
\end{table}

\subsection{User Experience}
\textit{August} required significantly fewer questions, with a mean of 16 questions, compared to symptom checkers that asked 29 questions, as shown in Table~\ref{tab:average_questions}.

\begin{table*}[htbp]
\centering
\caption{Average Questions Asked by System}
\label{tab:average_questions}
\begin{tabularx}{\textwidth}{@{}>{\raggedright\arraybackslash}p{0.35\textwidth}
                                >{\centering\arraybackslash}X
                                >{\centering\arraybackslash}X
                                >{\centering\arraybackslash}X@{}}
\toprule
\textbf{System} & \textbf{Avg Questions} & \textbf{Avg Questions} & \textbf{\% Fewer} \\
& \textbf{(Symptom Checker)} & \textbf{(August)} & \textbf{Questions} \\
\midrule
Cardiovascular            & 33 & 17 & 47\% \\
Dermatology               & 28 & 16 & 42\% \\
Endocrine                 & 29 & 18 & 38\% \\
ENT                       & 29 & 14 & 51\% \\
GI                        & 31 & 16 & 49\% \\
Hematology                & 29 & 16 & 44\% \\
Infectious                & 32 & 15 & 53\% \\
Nephrology                & 30 & 16 & 46\% \\
Neurology                 & 30 & 15 & 49\% \\
Obstetrics and Gynecology & 27 & 15 & 44\% \\
Ophthalmology             & 27 & 13 & 51\% \\
Orthopedics and Rheumatology & 29 & 16 & 45\% \\
Respiratory               & 31 & 15 & 52\% \\
Urology                   & 23 & 15 & 35\% \\
\midrule
\textbf{Grand Total}      & \textbf{29} & \textbf{16} & \textbf{47\%} \\
\bottomrule
\end{tabularx}
\end{table*}

\section{Discussion}
Our hope with this framework is to advance the state of the art in benchmarking diagnostic accuracies of Health AIs. We aim to improve every part of this system while highlighting the necessity of accounting for both subjective and objective information to provide a concise diagnosis. We aim to extend and make our patient actors available to the broader community who aim  to benchmark similar AIs. These benchmark results from August demonstrate a paradigm shift in how individuals interact with healthcare technology. \textit{August}, an advanced AI system, not only achieves high accuracy when evaluated against gold-standard clinical vignettes but also engages users in concise and empathetic conversations. By recommending appropriate courses of action tailored to each user's needs, \textit{August}'s performance surpasses that of traditional symptom checkers across multiple dimensions.

\subsection{\textit{August}'s Approach}
In contrast to the standard shortcomings indicated in Symptom Checkers, \textit{August} addresses these shortcomings by engaging users in meaningful conversations~\citep{ref46}. Its empathetic approach ensures that users feel heard and validated —essential for effective healthcare interactions. By processing both objective symptoms and subjective experiences, \textit{August} provides more accurate assessments and personalized recommendations. This holistic approach enhances the UX, increasing user trust and adherence to the provided guidance. A study comparing AI chatbot responses to physician answers to patient questions found that chatbot responses were preferred in the majority of evaluations, with chatbot responses rated significantly higher in both quality and empathy~\citep{ref43}. This suggests that AI assistants like \textit{August} could effectively aid clinicians by drafting responses to patient inquiries, potentially easing clinician workload and improving patient outcomes. Patients' medicine information needs significantly differ by disease type and age~\citep{ref34}, which is addressed by \textit{August}. Patient-centred decision-making (PCDM) significantly improves healthcare outcomes by adapting care plans to individual patient contexts~\citep{ref44}.

\subsection{\textit{August}'s Role in Addressing Healthcare Gaps}
\textit{August's} ability to deliver consistent and evidence-based information provides an advantage over human providers, who may vary in expertise and approach. This consistency enhances the overall quality of care and helps reduce disparities in healthcare delivery.\textit{August's} comprehensive patient histories provide a better opportunity to provide empathetic care based on the afflicted disease and are capable of identifying when specialist care is needed. This proactive identification can expedite referrals and ensure users receive appropriate care more quickly~\citep{ref45}. The constant accessibility provided via a digital medium also enhances healthcare access, allowing users from diverse economic and social backgrounds an opportunity to understand and seek help when needed while not incurring unnecessary expenses.  

\section{Strengths and Limitations}
The use of clinical vignettes has limitations in replicating the complexities of real patient encounters~\citep{ref53,ref54}. Vignettes are simplified representations that often fail to capture the nuances of individual patient presentations, including varying symptom descriptions, co-morbidities, and unpredictable patient behaviours. Real patients may present with atypical symptoms or fail to recognize certain signs, leading to diagnostic challenges that vignettes cannot replicate~\citep{ref55}. This limits representativeness, which can decrease the generalizability of findings to real-world clinical settings.
Having the patient actor communicate in grammatically correct English does not reflect the linguistic diversity encountered in clinical practice. Patients come from varied backgrounds and may have limited English proficiency, strong accents, use colloquial language, or exhibit speech impairments. These factors can impact communication clarity, leading to misunderstandings or requiring additional interpretive efforts by clinicians~\citep{ref56,ref57}. By not accounting for language barriers and communication challenges, the study may overlook potential obstacles faced by diagnostic tools and healthcare providers when accurately assessing patient information.
In reality, patients often lack explicit knowledge of which symptoms are relevant to their condition~\citep{ref58,ref59}. They may omit crucial details, provide vague descriptions, or focus on less relevant symptoms due to anxiety, misunderstanding, or forgetfulness. The patient actor's ability to accurately provide pertinent positives and negatives upon prompting does not mimic the exploratory nature of clinical interviews, where clinicians must skillfully guide conversations to elicit essential information. This can lead to an underestimation of the diagnostic tool's ability to handle incomplete or inaccurate patient inputs.
The absence of physical examination findings and diagnostic tests is a major limitation. Many conditions cannot be accurately diagnosed without these critical pieces of information. Physical signs observed during an examination, along with laboratory and imaging results, provide objective data that complement patient-reported symptoms~\citep{ref60}. By excluding these elements, the study reduces the diagnostic process to symptom analysis alone, which may not be sufficient for accurate diagnosis. This limitation also affects comparability between human clinicians, who rely on these data, and symptom checkers who cannot incorporate such information.
The vignette approach may inadvertently create homogenized patient presentations that do not reflect the variability seen in clinical practice. Patients with the same condition can present differently based on factors such as age, gender, ethnicity, and socioeconomic status. A lack of diversity in the vignettes may limit the applicability of the findings across different patient populations.

\section{Future Course of Action}
\subsection{Increase in Number and Spectrum of Vignettes}
\begin{itemize}
    \item \textbf{Expand Case Diversity}: Utilize a larger and more diverse set of clinical vignettes covering a wide range of medical conditions, severities, and presentations, including common illnesses, rare diseases, atypical presentations, and cases with comorbidities. Ensuring linguistic, sociocultural, and contextual diversity in the patient actor would be crucial.
\end{itemize}
\begin{itemize}
    \item \textbf{Incorporate Patient Demographics}: Include vignettes representing different ages, genders, ethnicities, and socioeconomic backgrounds to assess the diagnostic tool's performance across diverse patient groups. Conversational diagnostic AI, while enhancing clinical care, poses risks to protected groups due to historical biases in data, necessitating careful thought in model design and usage to ensure equitable health outcomes for all patients~\citep{ref49}.    
\end{itemize}

\subsection{Randomized Controlled Trial (RCT) with Real Human Patients}
\begin{itemize}
    \item \textbf{Conduct Gold-Standard Research}: Implementing an RCT involves testing the diagnostic tool in real clinical settings with actual patients, providing the highest level of evidence for its effectiveness.
\end{itemize}
\begin{itemize}
    \item \textbf{Include Physical Examination and Diagnostic Data}: RCTs allow for integrating physical examination findings, laboratory results, and radiological data, offering a comprehensive assessment of the diagnostic process.
\end{itemize}
\begin{itemize}
    \item \textbf{Assess Real-World Performance}: Testing with real patients captures the unpredictability and complexity of actual clinical encounters, including incomplete information, patient anxiety, and non-adherence to expected communication patterns.
\end{itemize}
\begin{itemize}
    \item \textbf{Evaluate Clinical Outcomes}: Measuring actual patient outcomes provides insights into the diagnostic tool's impact on clinical decision-making and patient care.
\end{itemize}

\section{Conclusion}
This study aims to be one of many steps toward creating accurate and safe health AI systems. Through this benchmarking framework, we hope to inspire others to establish the validity of their models. To facilitate this, we are committed to making it easier for anyone to benchmark their health AI platform.

The integration of AI-driven conversational agents, like August, represents a promising advancement in healthcare. These tools demonstrate the ability of health AI to engage users through natural language, accurately interpret patient-reported symptoms, and provide reliable diagnostic suggestions with notable precision. By complementing traditional diagnostic methods, health AIs like August have the potential to enhance patient engagement, expand access to medical guidance, and alleviate some of the burdens on healthcare systems.

Nevertheless, certain limitations, such as reliance on clinical vignettes and the absence of physical examinations, highlight areas for improvement. Integrating physical assessments into AI evaluations would significantly enhance the effectiveness of these tools. Rigorous testing in real-world clinical settings is crucial to fully understand August’s capabilities and ensure its safe integration into medical practice.

This approach leverages the rapid advancements in AI to address complex challenges in medicine, ultimately aiming to improve patient outcomes while preserving the essential human aspects of care.

\section{Author Contributions Statement}
D.B., and A.M. conceived the experiment(s). D.B. and S.A. conducted the experiment(s). D.B. and S.A. analyzed the results. S.A. wrote the manuscript. D.B., A.M., and S.A. reviewed the manuscript.

\section{Data Availability Statement}
Requests for data may be sent to contact@getbeyondhealth.com

\clearpage 
\appendix

\onecolumn
\section{Patient-AI Interaction Table}
\begin{longtable}{|p{0.45\textwidth}|p{0.45\textwidth}|}
\hline
\textbf{Patient Messages} & \textbf{Health AI} \\
\hline
\endfirsthead
\hline
\textbf{Patient Messages} & \textbf{Health AI} \\
\hline
\endhead
Hi August & Hello! I'm here to help you with any health concerns you might have. To get started, could you please tell me your age and sex? \\
\hline
Sure, I'm a 23 year old, male. & Thank you for sharing that. Are you currently taking any medications? \\
\hline
No, I'm not taking any medications right now. & Great, thanks for letting me know. Are you currently experiencing any symptoms or health concerns that you'd like to discuss? \\
\hline
Yes, I've been having this dull-aching pain in my shoulder for about a week now. It's not getting any better, and it stops me from moving my shoulder. & I'm sorry to hear about your shoulder pain. How did this pain start, and have you noticed any changes or events that might have triggered it? \\
\hline
I didn't have any recent trauma or injury. It just started about a week ago and hasn't improved. & Thank you for that information. How would you describe the pain? Is it constant, or does it come and go? And does anything make it better or worse? \\
\hline
The pain is a dull ache and it's pretty constant. It gets worse at night or if I touch or lie on my arm. & It sounds like the pain is persistent and worsens with certain activities or positions. Have you noticed any other symptoms, such as swelling, redness, warmth around the shoulder, or any numbness or tingling in your arm or hand? \\
\hline
No, I haven't noticed any swelling, redness, or numbness or tingling in my arm or hand. & Thank you for clarifying that. Have you experienced any recent changes in your physical activity or routine, such as starting a new exercise program or engaging in repetitive movements that might have affected your shoulder? \\
\hline
Yes, I'm a tennis player, and I've been preparing for a big tournament for about a month now. & Given your involvement in tennis and the recent increase in activity, it's possible that your shoulder pain could be related to overuse or strain. Have you experienced any similar shoulder issues in the past, or is this the first time you're dealing with this kind of pain? \\
\hline
This is the first time I'm dealing with this kind of pain. & Thank you for sharing that. Considering your tennis activity, it's important to explore the possibility of conditions like rotator cuff tendinitis or shoulder impingement, which are common in athletes. Have you noticed any weakness in your shoulder or difficulty performing specific movements, such as lifting your arm above your head? \\
\hline
I haven't noticed any weakness, but I do have difficulty putting my jacket on because of the pain. & The difficulty you're experiencing with certain movements, like putting on a jacket, aligns with shoulder impingement or rotator cuff issues. To further narrow down the cause, have you tried any treatments or remedies, such as rest, ice, or over-the-counter pain medications, and if so, have they provided any relief? \\
\hline
I haven't tried any treatments or remedies yet. & Since you haven't tried any treatments yet, it might be helpful to start with some basic self-care measures. However, before we discuss that, I want to ensure we rule out any other potential causes. Have you experienced any fever, fatigue, or other systemic symptoms that might suggest an infection or another underlying condition? \\
\hline
No, I haven't experienced any fever, fatigue, or other symptoms like that. & Thank you for confirming that. Given your symptoms and the context of increased tennis activity, it seems likely that your shoulder pain could be due to overuse, possibly leading to conditions like rotator cuff tendinitis or shoulder impingement. These are common in athletes who perform repetitive overhead motions.

To help manage the pain, you might consider:

\begin{itemize}[leftmargin=*]
    \item Resting the shoulder and avoiding activities that exacerbate the pain.
    \item Applying ice to the affected area to reduce inflammation.
    \item Using over-the-counter pain relievers, if needed, to manage discomfort.
\end{itemize}

If the pain persists or worsens, it would be wise to consult a healthcare professional for a more thorough evaluation and possibly physical therapy. They can provide exercises to strengthen the shoulder and improve flexibility, which can be beneficial in preventing future issues.

If there's anything else you'd like to discuss or if you have further questions, feel free to let me know! \\
\hline
Thank you for the advice. I think that's all I needed to know for now. I appreciate your help! & You're welcome! I'm glad I could assist you. If you have any more questions or need further advice in the future, don't hesitate to reach out. Take care and best of luck with your tournament preparation! \\
\hline
\end{longtable}
\twocolumn

\end{document}